\documentclass[runningheads]{llncs}
\usepackage{graphicx}

\usepackage[urlcolor=blue,colorlinks=true]{hyperref}

\usepackage{marvosym} 
\usepackage{amssymb} 
\usepackage{amsmath}
\usepackage{color} 
\definecolor{imginfocolor}{cmyk}{1,1,0,0}
\definecolor{pointcolor}{cmyk}{0,1,1,0}
\definecolor{probabilitymapcolor}{cmyk}{0,0,0,0.3}

\graphicspath{{img/}}

\begin{document}
\title{Learning Shape Representation on Sparse Point Clouds for Volumetric Image Segmentation}
\titlerunning{Learning Shape Representation on Sparse Point Clouds for Segmentation}
%
\author{Fabian Balsiger\inst{1}\textsuperscript{(\Letter)}\orcidID{0000-0001-7577-9870} \and
Yannick Soom\inst{1} \and
Olivier Scheidegger\inst{2} \and
Mauricio Reyes\inst{1}}

\authorrunning{F. Balsiger et al.}
%
\institute{Insel Data Science Center, Department of Neuroradiology, Inselspital,\\Bern University Hospital, University of Bern, Switzerland \\ \email{fabian.balsiger@artorg.unibe.ch} \and
Support Center for Advanced Neuroimaging (SCAN), Institute for Diagnostic and Interventional Neuroradiology, Inselspital, Bern University Hospital,\\University of Bern, Switzerland}

\maketitle              
\begin{abstract}
Volumetric image segmentation with convolutional neural networks (CNNs) encounters several challenges, which are specific to medical images. Among these challenges are large volumes of interest, high class imbalances, and difficulties in learning shape representations. To tackle these challenges, we propose to improve over traditional CNN-based volumetric image segmentation through point-wise classification of point clouds. The sparsity of point clouds allows processing of entire image volumes, balancing highly imbalanced segmentation problems, and explicitly learning an anatomical shape. We build upon PointCNN, a neural network proposed to process point clouds, and propose here to jointly encode shape and volumetric information within the point cloud in a compact and computationally effective manner. We demonstrate how this approach can then be used to refine CNN-based segmentation, which yields significantly improved results in our experiments on the difficult task of peripheral nerve segmentation from magnetic resonance neurography images. By synthetic experiments, we further show the capability of our approach in learning an explicit anatomical shape representation.

\end{abstract}

\setcounter{footnote}{0} 

\section{Introduction}
Convolutional neural networks (CNNs) have enabled significant progress in medical image segmentation. Despite this progress, CNN-based segmentation still faces several challenges to process large volumes of interest. Among these challenges: difficulty to process target structures that span entire or large image volumes along with limited memory of graphics processing units (GPUs), limited image information between adjacent image slices due to highly anisotropic image resolutions (e.g., large image slice thicknesses and gaps), and highly imbalanced problems due to sparse target structures. Moreover, these challenges also hinder learning shape representations, which might be desirable for segmentation and especially for target structures with distinct anatomy~\cite{Ravishankar2017,Oktay2018,Dalca2018}.

These challenges have been addressed in a rather separate manner~\cite{Litjens2017}. To tackle large volumes of interest and limited GPU memory, 2.5-D, dual-pathway, or patch-based processing has been proposed. Similarly, handling of anisotropic image resolution has been performed by only using a few image slices or entirely rely on slice-wise processing. Finally, dedicated loss functions were introduced to cope with highly imbalanced problems. An example with the aforementioned challenges existing is the segmentation of peripheral nerves from magnetic resonance neurography (MRN) images~\cite{Balsiger2018a}. First, peripheral nerves span the entire image volume from the most proximal to most distal image slice. Second, they are hard to distinguish from muscular tissue, which hinders patch-based processing and calls for strategies incorporating global context information. Third, a large image slice thickness (4.4~mm) of the MRN images hinders 2.5-D and dual-pathway processing. Fourth, the problem is highly imbalanced with peripheral nerves on average only accounting for 0.14~\% of the voxels in the image volume. Lastly, peripheral nerves have a distinct tubular-like anatomical shape, which make them an excellent target for shape learning.

\begin{figure}
\centering
\includegraphics[width=.8\textwidth]{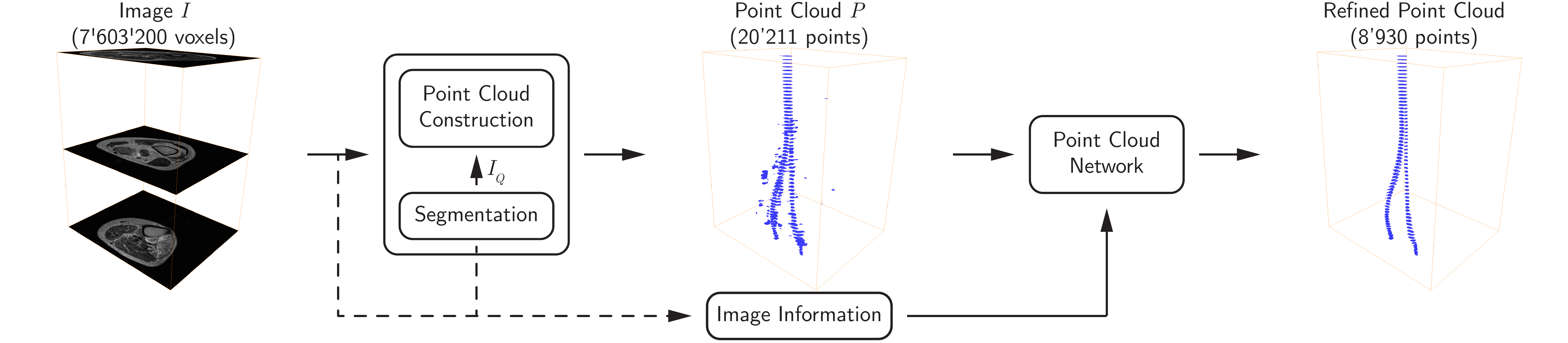}
\caption{Overview of the proposed approach. A volumetric image $I$ is processed by a classifier to obtain a probability map $I_Q$. We then transform $I_Q$ into the sparse representation of a point cloud $P$. The point cloud is processed by a point cloud network that classifies each point into foreground or background yielding a refined segmentation. The network incorporates not only Cartesian point coordinates but also image information around each point. In this work, we obtained image information from $I_Q$.}
\label{fig1}
\end{figure}

We propose to improve over traditional CNN-based volumetric image segmentation using the representation of a point cloud to tackle the aforementioned challenges. In particular, we refine a CNN-based segmentation by transforming the problem of volumetric image segmentation into a point cloud segmentation, wherein a voxel-wise classification becomes a point-wise classification (Fig.~\ref{fig1}). This has several advantages: i) point clouds are a more efficient way to represent sparse anatomical shapes than voxel-wise representations, ii) the sparsity of point clouds allows processing entire image volumes at once and efficiently leveraging volumetric information, iii) high class imbalances can be significantly reduced, and iv) it allows learning anatomical shape explicitly. For instance, the right-most point cloud in Fig.~\ref{fig1} shows the compact representation of the entire structure with only 8930 points (0.12~\% of the voxels) and an almost balanced classification problem in the point cloud with 20211 points (i.e., from original 0.1/99.9 to 44/56 class ratio). We further propose to enrich point cloud information with image information extracted around each point, which enables us to maintain a compact model while jointly leveraging image information. As use case, we evaluated our approach to peripheral nerve segmentation from MRN images of the thigh, and show that the proposed approach can improve significantly over CNN-based segmentation. By synthetic experiments, we further show that our approach is capable of explicitly learning and exploiting the tubular-like shape of peripheral nerves for image segmentation.

\section{Methodology}

\subsection{Point Cloud Construction}
In a first step, we transform a volumetric image $I$ into a point cloud $P = [\textbf{p}_1, \textbf{p}_2, \dots, \textbf{p}_N]$ with $N$ points $\textbf{p}_i \in \mathbb{R}^3$ as follows: First, we obtain a probability map $I_Q$ from a classifier trained to segment the target structure (CNN for peripheral nerves in our experiments). We threshold $I_Q$ at $\theta$ to get the point cloud representation $P$, i.e., every voxel $\textbf{v} \in I_Q$ with a probability $q \in [0, 1]$ larger than $\theta$ is a point $\textbf{p}_i = (x, y, z)$ with the Cartesian coordinates of $\textbf{v}$. Therefore, a point cloud $P$ consists of a different amount of points depending on the target structure's size as well as the classifier's segmentation confidence. We also remark, that no point correspondence is needed by the point cloud network.

\subsection{Point Cloud Network}
\begin{figure}
\centering
\includegraphics[width=0.8\textwidth]{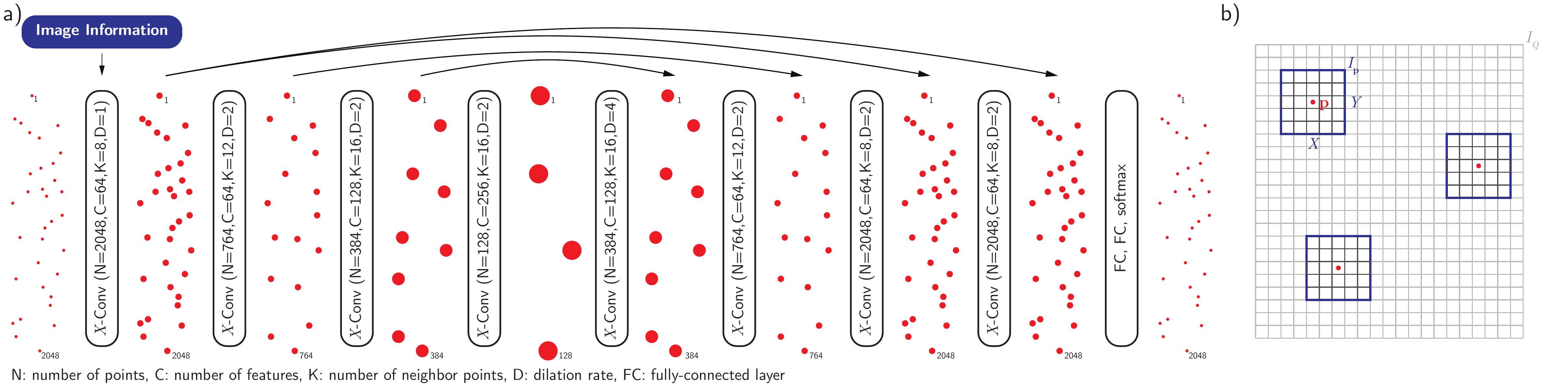}
\caption{a) The proposed network architecture, which takes a point cloud and image information as input. The output is the probability of a point belonging to the target structure. The size of the dots \textcolor{pointcolor}{$\bullet$} illustrates the number of features. b) Illustration of the formation of the image information. We extract features from a volume of interest $I_\textbf{p}$ (\textcolor{imginfocolor}{blue}) from $I_Q$ (\textcolor{probabilitymapcolor}{light gray}, probability values not shown for the sake of clarity) around each point $\textbf{p}$ (\textcolor{pointcolor}{$\bullet$}) in a point cloud, which yields the image information.}
\label{fig2}
\end{figure}

The point cloud network builds upon PointCNN introduced by Li et al.~\cite{Li2018c}. PointCNN follows the well-known encoding-decoding structure that gradually downsamples the point cloud to capture context followed by upsampling the point cloud and combining features through skip connections (Fig.~\ref{fig2}a). The core of PointCNN is the $\mathcal{X}$-Conv operator, which is the counterpart of the convolution operator for unstructured data. At each encoding step, an input point cloud $P_{in}$ is reduced to a set $P_{out} \subset P_{in}$ of $N = \vert P_{out} \vert \leq \vert P_{in} \vert$ representative points $\textbf{p}$ using farthest point sampling. At each representative point $\textbf{p}$, the $\mathcal{X}$-Conv operator extracts $C$ features from the $K$ nearest neighbor points of $\textbf{p}$ in $P_1$ yielding a feature rich representative point $\textbf{p}$. $K$ can be seen as the receptive field of the network, which is increased towards the bottleneck by a dilation rate $D$, arriving ultimately at a receptive field of $D * K$. A decoding step works similarly, with the difference that $P_{out}$ now has more points and fewer features compared to $P_{in}$ ($\vert P_{out} \vert \geq \vert P_{in} \vert$). Further, the point clouds of the encoding path are concatenated following the principle of skip connections. After the last $\mathcal{X}$-Conv operator, two fully-connected (FC) layers reduce the features of each point to the number of classes. Finally, a softmax function assigns the class probabilities to each point.\footnote{We remark that other point cloud architectures might also be feasible for the task at hand. Experiments with PointNet and PointNet++~\cite{Qi2017}, two popular architectures and pioneers in deep learning-based point cloud processing, showed slightly worse performance and are omitted here for clarity.}

Despite the capability of PointCNN to reason on point clouds, it might be beneficial to jointly leverage the rich information contained in the volumetric images together with the Cartesian point coordinates. Inspired by~\cite{Rempfler2016b}, we extract image information from the probability map $I_Q$, which gives a strong indication of whether a point belongs to the target structure or not based on the raw image intensities. In particular, we define the image information for each point $\textbf{p}$ to be features extracted from a volume of interest $I_\textbf{p} \in \mathbb{R}^{X \times Y \times Z}$ centered at a point $\textbf{p}$ (Fig.~\ref{fig2}b). $I_\textbf{p}$ is processed by a feature extraction module, which consists of a sequence of two 3-D convolutions (kernel size of 3 and stride of 1, 4 and 8 channels, respectively) with ReLU activation function, batch normalization, and max pooling operation. The features are then reshaped to a vector of size 64 and fed into the first $\mathcal{X}$-Conv operator together with the point's Cartesian coordinates. The feature extraction module and PointCNN can be trained end-to-end. We set $X = Y = Z = 5$ in our experiments. For reproducibility, the code is publicly available.\footnote{\href{https://github.com/fabianbalsiger/point-cloud-segmentation-miccai2019}{https://github.com/fabianbalsiger/point-cloud-segmentation-miccai2019}}

\section{Experiments and Results}

\subsection{Data and Baselines}
\label{sec:data-baseline}
We used 52 MRN images of the thigh of healthy volunteers (n=10) and patients (n=42) to evaluate our method. The sciatic nerve has been manually segmented by three clinical raters and these manual segmentations have been merged using voxel-wise majority voting to obtain a consensus ground truth, to which all results were compared. In a four-fold cross-validation, a CNN was trained, as in~\cite{Balsiger2018a}, to segment the peripheral nerves using the consensus ground truth. This segmentation, termed \texttt{CNN}, served as baseline as well as for the construction of the point cloud $P$ at threshold $\theta = 0.1$. As a second baseline \texttt{CNN-P}, we compared against a cascaded CNN~\cite{Havaei2017} that uses $I$ and the probability map $I_Q$ of the baseline \texttt{CNN} as inputs, which simulates the use of the proposed image information originating from $I_Q$. We followed the same training procedure as for \texttt{CNN}. From the three rater segmentations, two human rater variabilities were additionally obtained for comparison: a rater-to-rater variability (\texttt{R-R}) and a rater-to-consensus ground truth variability (\texttt{R-GT}). We remark that \texttt{R-R} overestimates and \texttt{R-GT} tends to underestimate the true rater variability.

\subsection{Network Training}
\label{sec:training}
We pre-processed the Cartesian coordinates of each $P$ to lie in the unit cube $[-1, 1]^3$ and $I_Q$ for extracting the image information to zero mean and unit variance on a subject level. We trained our network with $N = 2048$ input points. Therefore, during the training we randomly split a point cloud $P$ into smaller point clouds with 2048 points. This extraction randomly permuted the point clouds during the training, and we added random 3-D rotations and jittering of the point clouds as additional data augmentation. A batch consisted of eight point clouds, and we trained the network for $40$ epochs using Adam optimizer with a learning rate of $0.01$ and cross-entropy loss. The training followed the same four-fold cross-validation as in Sec.~\ref{sec:data-baseline} and hyperparameters were selected only on one fold. Note that during testing, we also randomly extract subsets of 2048 points until all points of a point cloud have been classified. We repeated this process ten times and did a majority voting to obtain the final point classification, which yielded slightly more robust classifications.

\subsection{Ablation Study}
\begin{figure}
\centering
\includegraphics[width=0.7\textwidth]{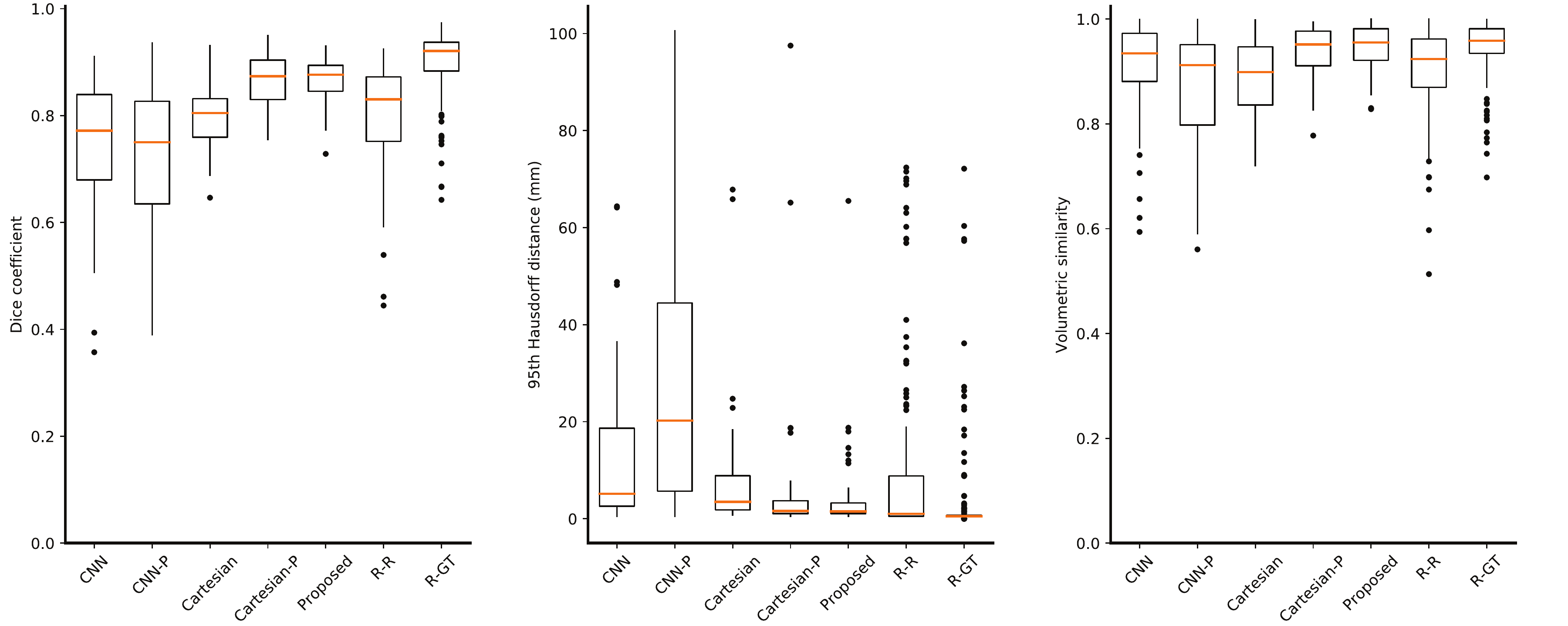}
\caption{Box-and-whisker plots for the metrics Dice coefficient, 95\textsuperscript{th} Hausdorff distance, and volumetric similarity.}
\label{fig3}
\end{figure}

To study the benefits of the proposed approach, we evaluated three variants: point cloud with Cartesian coordinates only (\texttt{Cartesian}, equivalent to PointCNN only), point cloud with probability $q$ (\texttt{Cartesian-P}), i.e., input is $(x, y, z, q)$ similar to setting $X = Y = Z = 1$, and point cloud with image information from a $X = Y = Z = 5$ volume of interest (\texttt{Proposed}). \texttt{Cartesian-P} was added to show the benefit of neighborhood information around each point. Fig.~\ref{fig3} shows box-and-whisker plots for the three metrics Dice coefficient (DICE), 95\textsuperscript{th} Hausdorff distance (HD95), and volumetric similarity (VS). All point cloud-based variants outperform the CNN-based segmentations (\texttt{CNN} and \texttt{CNN-P}). With image information, our approach achieves statistically significant better results compared to the \texttt{R-R} variability for the DICE and the VS (both p $\leq$ 0.001) and on-par results for the HD95 (p $=$ 0.100). Regarding \texttt{R-GT} variability, we achieve on-par performance for the VS (0.946 $\pm$ 0.041 vs. 0.944 $\pm$ 0.055, p $=$ 0.600), and slightly decreased performances for the DICE (0.866 $\pm$ 0.044 vs. 0.899 $\pm$ 0.061, p $\leq$ 0.001) and the HD95 (4.5 $\pm$ 9.7~mm vs. 3.9 $\pm$ 11.1~mm, p $\leq$ 0.001). Compared to \texttt{Cartesian-P}, the \texttt{Proposed} image information performs slightly better, particularly for the HD95 with 4.5 $\pm$ 9.7~mm vs. 6.1 $\pm$ 16.1~mm (p $=$ 0.57), although not statistically significant. Statistical tests with a Mann-Whitney U test and a significance level of 0.05.

\begin{figure}
\centering
\includegraphics[width=0.6\textwidth]{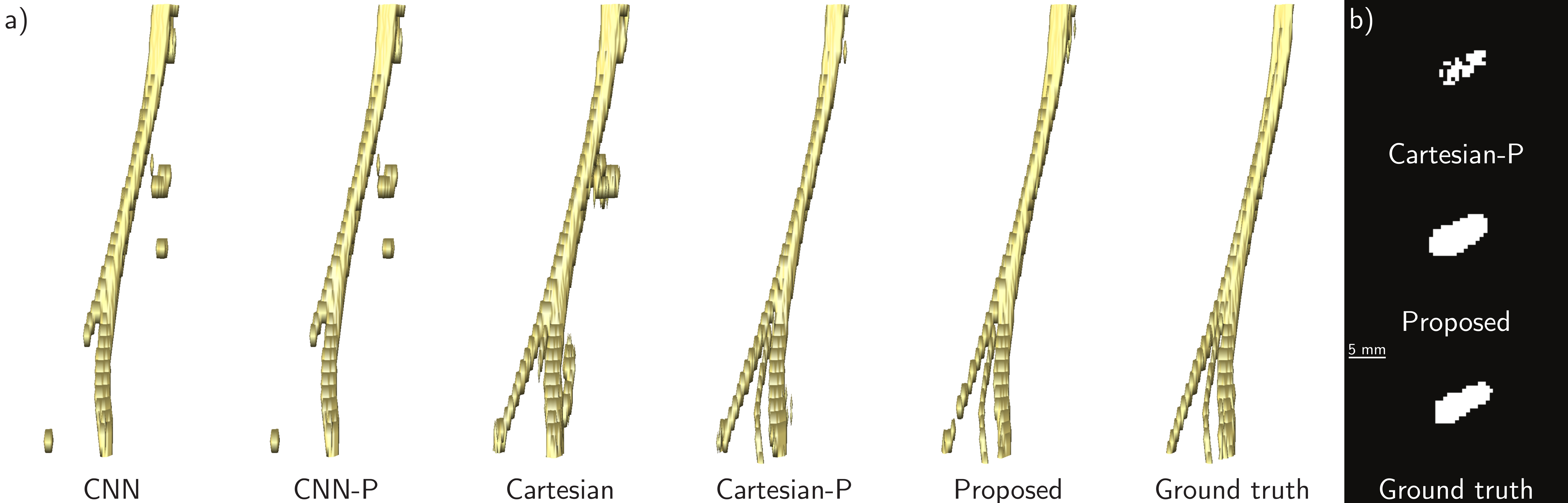}
\caption{Qualitative results: a) 3-D renderings of segmentation results. b) Worse in-plane segmentation results occurred with higher frequency for \texttt{Cartesian-P} than \texttt{Proposed}.}
\label{fig4}
\end{figure}

The 3-D renderings in Fig.~\ref{fig4}a illustrate the improvement of the final segmentation by the proposed method compared to CNN-based segmentation, \texttt{Cartesian} and \texttt{Cartesian-P}. The differences between \texttt{Cartesian-P} and \texttt{Proposed} are negligible in 3-D but accentuate in 2-D (Fig.~\ref{fig4}b). In some few cases, a high HD95 was found mostly due to misclassification of points arranged in a tubular-like way in combination with image information of high probability, i.e. also misclassified by the CNN-based segmentation. We, therefore, investigated the sensitivity to shape and image information with a synthetic experiment, described below.

\subsection{Sensitivity to Shape and Image Information}
We conducted an experiment to confirm the hypothesis that our point cloud processing is capable of learning an explicit shape representation and is not only dependent on the probabilities of $I_Q$. We created two synthetic cases, one resembling a straight sciatic nerve and the other resembling a sciatic nerve with branching into tibial and fibular nerves. The nerves were manually drawn in ITK-SNAP\footnote{\url{http://www.itksnap.org/}} with a circular in-plane shape (diameter of ten voxels, based on the size of peripheral nerves in our data). For each case, we then assigned the probabilities $q = \{0.1, 0.3, 0.5, 0.7, 0.9\}$ to the synthetic nerves, which we additionally smoothed with a Gaussian filter ($\sigma = 1$) to simulate less confident boundaries arriving at a synthetic probability map $I_Q$. We then classified the cases using the proposed approach, which was trained on real data. Independent of the case or the probability, our approach consistently classified the point clouds correctly as peripheral nerves with Dice coefficients of $0.953 \pm 0.015$.

\begin{figure}
\centering
\includegraphics[width=0.7\textwidth]{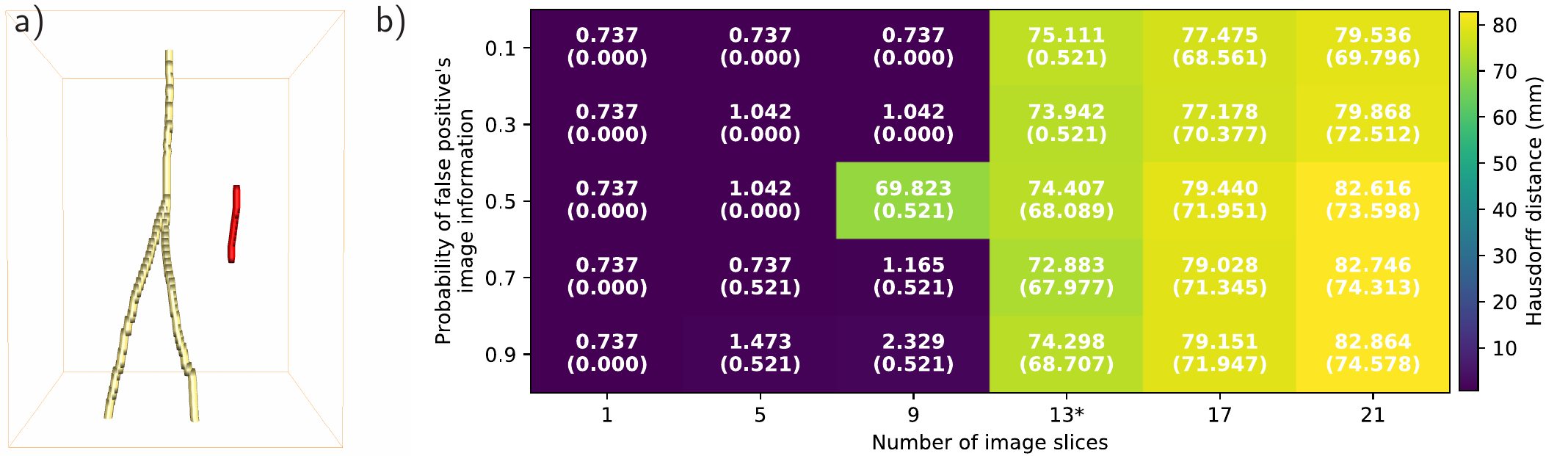}
\caption{Second synthetic experiment: a) 3-D rendering of the synthetic sciatic nerve (in yellow) with a tubular-like false positive that spans 13 image slices (in red), b) heat map of the Hausdorff distance (95\textsuperscript{th} Hausdorff distance in parenthesis) after refinement depending on the image information's probability and size of the false positive. The asterisk indicates the false positive shown in a).}
\label{fig5}
\end{figure}

In a second experiment, we then investigated the influence of the shape and image information on false positive removal. Inspired by a case in our data, where a vein was misclassified, we manually draw a tubular-like false positive spanning from 1 up to 21 image slices (Fig.~\ref{fig5}a shows a synthetic nerve with a false positive spanning 13 image slices). Intuitively, our approach should remove the false positive when it only spans a few image slices, independent of the image information's probability, because the shape does not resemble a peripheral nerve. Therefore, we varied the probability of the false positive ($q = \{0.1, 0.3, 0.5, 0.7, 0.9\}$) while fixing the probability of the synthetic nerve to be $0.5$. The heat map in Fig.~\ref{fig5}b shows that false positives spanning nine or fewer image slices get almost correctly removed independently of the image information's probability. Therefore, we concluded that the point cloud network learns a coarser anatomical shape resembling a peripheral nerve. It seems that shape is more important than the image information, which might only give an additional clue whether a certain point, e.g. at the boundary of the nerves, is classified as peripheral nerve or not.

\section{Discussion and Conclusion}
We used point clouds combined with image information to improve volumetric image segmentation based on common challenges in CNN-based segmentation. Our approach comes with the benefit of processing entire image volumes by leveraging the sparsity of a point cloud representation. Further, we showed that the point clouds allow us to explicitly learn the anatomical shape of the target anatomical structure. Jointly incorporating image information additionally improved the results on a coarse and local level (cf. Fig.~\ref{fig4}). During the development, we also experimented with using the raw image intensities or Hessian matrix entries, inspired by~\cite{Frangi1998}, as image information. However, none of them improved the results compared to directly using $I_Q$. We hypothesize that this might be attributed to the rich information encoded in $I_Q$. But theoretically, any kind of image information can be incorporated into the network using the proposed scheme, which leaves room for tailoring the image information to specific applications. Finally, we remark that the quality of the point cloud depends on the quality of $I_Q$. This is a limitation of the presented approach. Selecting the threshold $\theta$ is a trade-off between including false positives and false negatives, and while our first results did not show high sensitivity to $\theta$, a dedicated sensitivity analysis is subject to further work. 

In regard to the presented application of peripheral nerve segmentation, we achieved promising results that even approach rater to consensus ground truth variability (\texttt{R-GT}). This variability is known to be difficult to achieve because every rater contributes itself to the consensus ground truth. For the Dice coefficient, it might be almost impossible to match \texttt{R-GT} due to the sensitivity of the Dice coefficient to small structures. As a next step, the results need to be confirmed on other anatomical regions than the thigh, which might give further insights into learning shape representations.

We think that the proposed transformation to point clouds might be applicable to other anatomical structures, where the conditions of sparse anatomical structure, i.e. the point cloud represents a more compact representation than a volumetric image, and distinct anatomical shape are present. Such anatomical structures could, for instance, be the vascular system (e.g., aorta segmentation) or the pulmonary system (e.g., airway tree segmentation). And as mentioned, the image information leaves room for tailoring to new applications.

In conclusion, we investigated using point clouds to improve volumetric medical image segmentation. By using sparse point clouds, combined with the proposed image information, our approach can reason over a coarser anatomical shape, which leads to significantly improved segmentation results.

\subsubsection{\ackname}
This research was supported by the Swiss National Science Foundation (SNSF). The authors thank the NVIDIA Corporation for their GPU donation and Alain Jungo for fruitful discussions.

\bibliographystyle{splncs04}
\bibliography{library_short}

\begin{thebibliography}{10}
\providecommand{\url}[1]{\texttt{#1}}
\providecommand{\urlprefix}{URL }
\providecommand{\doi}[1]{https://doi.org/#1}

\bibitem{Balsiger2018a}
Balsiger, F., et~al.: {Segmentation of Peripheral Nerves from Magnetic
  Resonance Neurography: A Fully-automatic, Deep Learning-based Approach}.
  Frontiers in Neurology  \textbf{9}, ~777 (2018).
  \doi{10.3389/fneur.2018.00777}

\bibitem{Dalca2018}
Dalca, A.V., et~al.: {Anatomical Priors in Convolutional Networks for
  Unsupervised Biomedical Segmentation}. In: CVPR. pp. 9290--9299 (2018)

\bibitem{Frangi1998}
Frangi, A.F., et~al.: {Multiscale vessel enhancement filtering}. In: MICCAI,
  pp. 130--137. Springer (1998). \doi{10.1007/BFb0056195}

\bibitem{Havaei2017}
Havaei, M., et~al.: {Brain tumor segmentation with Deep Neural Networks}.
  Medical Image Analysis  \textbf{35},  18--31 (2017).
  \doi{10.1016/j.media.2016.05.004}

\bibitem{Li2018c}
Li, Y., et~al.: {PointCNN: Convolution On X-Transformed Points}. In: NIPS 31,
  pp. 828--838. Curran Associates (2018)

\bibitem{Litjens2017}
Litjens, G., et~al.: {A survey on deep learning in medical image analysis}.
  Medical Image Analysis  \textbf{42},  60--88 (2017).
  \doi{10.1016/j.media.2017.07.005}

\bibitem{Oktay2018}
Oktay, O., et~al.: {Anatomically Constrained Neural Networks (ACNNs)}. IEEE
  Transactions on Medical Imaging  \textbf{37}(2),  384--395 (2018).
  \doi{10.1109/TMI.2017.2743464}

\bibitem{Qi2017}
Qi, C.R., et~al.: {PointNet++: Deep Hierarchical Feature Learning on Point Sets
  in a Metric Space}. In: NIPS 30, pp. 5099--5108. Curran Associates (2017)

\bibitem{Ravishankar2017}
Ravishankar, H., et~al.: {Learning and Incorporating Shape Models for Semantic
  Segmentation}. In: MICCAI, Lecture Notes in Computer Science, vol. 10433, pp.
  203--211. Springer, Cham (2017). \doi{10.1007/978-3-319-66182-7\_24}

\bibitem{Rempfler2016b}
Rempfler, M., et~al.: {Reconstructing cerebrovascular networks under local
  physiological constraints by integer programming}. Medical Image Analysis
  \textbf{25}(1),  86--94 (2016). \doi{10.1016/j.media.2015.03.008}

\end{thebibliography}

\end{document}